\documentclass[letterpaper, 10 pt, conference]{ieeeconf}  

\IEEEoverridecommandlockouts                              

\usepackage{graphicx} 
\usepackage{float}
\usepackage[T1]{fontenc}
\usepackage{cite}
\usepackage{listings}
\title{\LARGE \bf
LLMs for Robotic Object Disambiguation
}

\author{Connie Jiang$^{1}$, Yiqing Xu$^{2}$, and David Hsu$^{2}$
\thanks{*This work was not supported by any organization}
\thanks{$^{1}$Massachusetts Institute of Technology}%
\thanks{$^{2}$Authors are with the Smart Systems Institute, National University of Singapore. 
{\tt\small conniej@mit.edu, yiqing.xu@u.nus.edu, dyhsu@comp.nus.edu.sg}}%
}

\begin{document}

\maketitle
\thispagestyle{empty}
\pagestyle{empty}

\begin{abstract}
The advantages of pre-trained large language models (LLMs) are apparent in a variety of language processing tasks. But can a LM’s knowledge be further harnessed to effectively disambiguate objects and navigate decision-making challenges within the realm of robotics? Our study reveals the LLM's aptitude for solving complex decision making challenges that are often previously modeled by Partially Observable Markov Decision Processes (POMDPs). A pivotal focus of our research is the object disambiguation capability of LLMs. We detail the integration of an LLM into an tabletop environment disambiguation task, a decision making problem where the robot’s task is to discern and retrieve a user’s desired object from an arbitrarily large and complex cluster of objects. Despite multiple query attempts with zero-shot prompt engineering (details can be found in the Appendix), the LLM struggled to inquire about features not explicitly provided in the scene description. In response, we have developed a few-shot prompt engineering system to improve the LLM’s ability to pose disambiguating queries. The result is a model capable of both using given features when they are available and inferring new relevant features when necessary, to successfully generate and navigate down a precise decision tree to the correct object—even when faced with identical options.

\end{abstract}

\section{Introduction}

Pre-trained large language models (LLMs) have demonstrated notable advantages in a variety of tasks, raising the question of whether their capabilities can extend to decision-making challenges in robotics~\cite{singh2023progprompt, li2022pre, chen2023introspective, hagendorff2023human, wang2019rat, kitaev2018multilingual, zhu2020incorporating, brown2020language, keskar2019ctrl}. Our research centers on the object disambiguation capability of LLMs, particularly within the context of tabletop environments. The task is to discern and retrieve a user’s desired object from a complex cluster of objects that can be in any arbitrary arrangement. 

The seemingly simple task of disambiguating an object from a scene actually involves several challenges:
\begin{itemize}
    \item \textbf{Developing a multi-step plan for disambiguation}: The model must generate a sequence of questions aimed at gathering relevant feature information. In other words, the questions should be designed in a way that maximizes the information gained at each step and lead to successful disambiguation with as few queries as possible. 
    \item \textbf{Inferring new features}: If the scene description provided is insufficient in providing enough features to uniquely disambiguate the object, the model must be able to deduce potential relevant features that were not ever previously mentioned in the description.
\end{itemize}

Past methods of solving this task tend to either enumerate all possible candidates \cite{zhang2021invigorate} or use a manually defined set of features to optimize information gain over time \cite{yang2022interactive}. However, these approaches have limitations. Enumerating every possible candidate is extremely \textit{inefficient}, while the use of predefined features confines the model's capability to recognizing only these known features, potentially resulting in \textit{incomplete} solutions.

Our method overcomes the \textit{inefficiency} and \textit{incompleteness} challenges and is capable of efficiently disambiguating any object from any arbitrarily large tabletop scene by harnessing the ``common sense'' of pre-trained LLMs with few-shot prompt engineering. Large language models are complete, in the sense that they are not limited to pre-set features. Rather, they are capable of using features that are either \textbf{explicitly stated} in the given scene description, or \textbf{inferred if stated features are not sufficient}. LLMs also are significantly more efficient than enumeration, as they are capable of categorizing features to ask more general and effective queries.

We assess our model using two primary metrics: the count of questions needed for successful disambiguation, which indicates \textit{efficiency}, and a success rate, which reflects \textit{completeness}. To better comprehend the improvement our method offers, we have set up four comparative benchmarks. These include the optimal feature split, human-like disambiguation reasoning, an enumeration approach (commonly used in many robotic systems like INVIGORATE \cite{zhang2021invigorate}), and the latest POMDP-ATTR, which disambiguates strictly based on color and location.


Our findings demonstrate the LLM’s capability to navigate decision trees and accurately identify the correct object, even when faced with identical options. Experimental results show an accuracy of \textbf{95.79\%} and significant improvement from enumeration, human performance, and POMDP-ATTR.

\begin{figure*}[]
    \centering
    \includegraphics[scale=0.7]{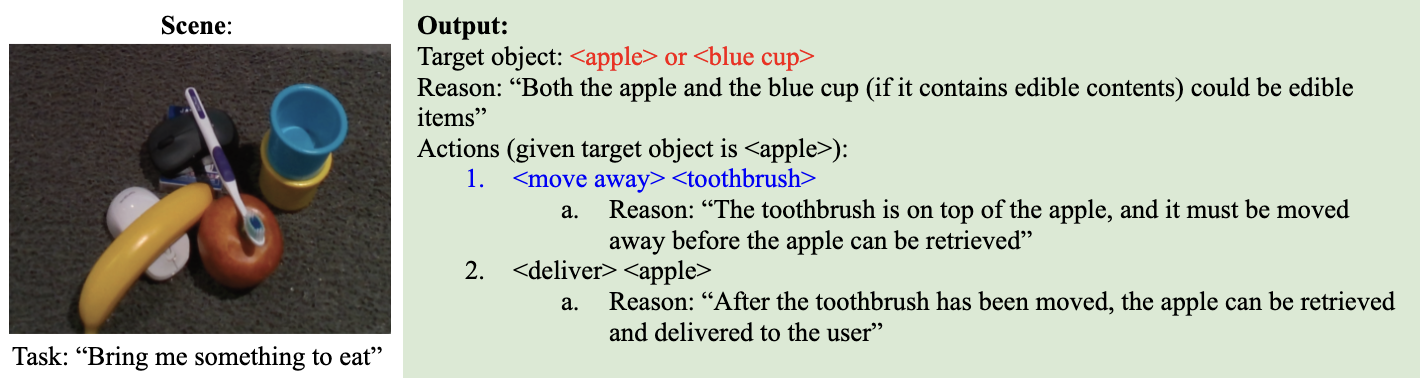} 
    \caption{\textbf{Generalization (red)}: The task given by the user was ``Give me something to eat'', and thus the user task specification has been generalized (either the apple or the contents inside the blue cup could be eaten, matching the task given by the user). \textbf{Occluding Objects (blue)}: the handling of occluding objects (the toothbrush is in the way of the apple, and must be removed first before the apple can be retrieved).}
    \label{fig:1}
\end{figure*}

\section{Related Work}

Disambiguation is a key task within the scope of robotics and has been approached using several methods in the past \cite{zhang2021invigorate, li2017learning, shridhar2020ingress, yang2022interactive, lutkebohle2009curious}. Meanwhile, the use of LLMs in robotics is a growing field of study \cite{mai2023llm, ding2023task, inoue2022prompter, singh2023progprompt, liu2023llm+}. This paper combines these two areas of research.

Regarding the task of disambiguation, the most common approach in robotic systems is currently enumeration, as evidenced in INVIGORATE and INGRESS-POMDP~\cite{zhang2021invigorate, shridhar2020ingress}. In these systems, the robot points individually to potential target objects (chosen either heuristically or with a POMDP) and poses a query along the lines of ``Is this the desired object?'' This method, while effective in reducing ambiguity, is not efficient, as each query only eliminates a single potential object instead of an entire class of objects. 

An alternative approach that has been explored is a greedy one, where the most ambiguous aspect or the most immediately discriminative question is addressed~\cite{tellexll2013toward, li2017learning, thomason2019improving}. For example, when differentiating between a green and a red apple, a question like “What color is your object?” would elicit a more informative response than “What fruit is your object?” This method is adept at generating relevant questions but fails to produce sequential queries, and thus fails in complicated scenes with numerous objects where more than one query is required for complete disambiguation.

The most recent method is attribute-guided disambiguation. The POMDP-ATTR model can disambiguate between objects by referring to specific attributes~\cite{yang2022interactive}. The fallback is that each attribute needs to be individually incorporated into the complex POMDP. Presently, POMDP-Attr model is limited in attributes to basic colors and the relative location between two objects. In their observation model, in order to ask a disambiguation question, they perform vocabulary-level sampling specific to color and location to form a query ~\cite{yang2021attribute}. The model’s limited feature scope makes it unable to infer relevant features that it has not previously learned. 

As we consider these limitations of prior methods, we can pivot our attention to the growing research field of LLMs in robotics~\cite{mai2023llm, ding2023task, inoue2022prompter, singh2023progprompt, liu2023llm+, silver2022pddl}. LLMs have proven capabilities in handling language processing tasks \cite{yang2023harnessing, naveed2023comprehensive, yin2023large}, but the more recent focus of interest is their use for decision making in robotics ~\cite{li2022pre}. The most notable applications of LLMs for interactive decision making have been centered around sequential decision-making tasks, which attempt to generate optimal paths to certain objectives. Further exploration has proved LLMs to be capable as zero-shot planners~\cite{huang2022language}. 

Back to the task of disambiguation, a task which requires human-robot interaction in a language-centered context, LLMs offer significant advantages. LLMs have already been proven to be capable as zero-shot planners \cite{zhang2023large, kojima2022large}, and with few-shot prompt engineering, their performance improves even more. They are not constrained by the earlier stated limitations (limited feature scope, the need for feature-specific vocabulary sampling, or exhaustive enumeration). Our model demonstrates a pre-trained LLM can efficiently disambiguate objects in an arbitrarily complex scene, inferring relevant features and positing disambiguation queries as needed. 

\begin{figure}[]
    \centering
    \includegraphics[scale=0.5]{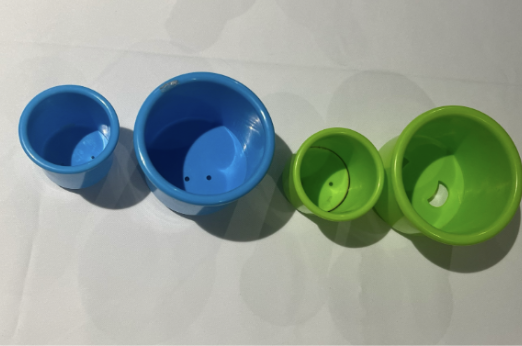}
    \caption{\textbf{Simple Scene Example with Zero-Shot Prompting} ``There are four cups in a line. Two are blue and two are green. They are of different sizes.'' In response, the model with zero-shot prompting proposes an action planner with two sequential questions, successfully completing the disambiguation task. The first seeks to understand the user's color preference (<ask> <``Do you prefer a certain color for the cup?''>), and the second determines the size of the target object (<ask> <``Do you want a large cup or a small one?''>).}
    \label{fig:2}
    \end{figure}

\section{Problem Formulation}

In this section, we aim to illustrate the variety of relevant tasks an LLM is capable of in a robotic task, encompassing several interrelated tasks relevant to the overall task pipeline. Our in-depth exploration, however, will be specifically on the \textbf{disambiguation task} detailed in subsection C. It is noteworthy that this particular question is where we have conducted experiments to validate and quantify our method's performance. We aim to show the efficiency and completeness of using few-shot prompt engineering for the decision making task of disambiguation.

\subsection{Generalizing User Requests} One immediate improvement that can be integrated into the platform pertains to the format of user inquiries. Traditionally, object retrieval functionality has been confined to handling requests for predefined objects (for instance, “Pass me the apple” or “Pass me the blue pen”). However, introducing an LLM provides a unique opportunity to significantly broaden this scope and personalize inquiries~\cite{wu2023tidybot}. Now, the task can be adapted to interpret and respond to any reasonable, generalized request (for instance, “Get me something to eat” or “Get me something to write with”). This development not only increases the utility of the platform but also demonstrates the transformative potential of LLMs in enriching human-robot interaction. Now, it is capable of interpreting broad, nonspecific prompts, making the robotic system more user-friendly and giving it a higher level of responsiveness and adaptability. 

\subsection{Maneuvering Occluding Objects} After identifying a target object, the next challenge is its successful delivery. This involves the non-trivial task of first relocating any objects that occlude the desired one. \cite{Danielczuk2019MechanicalSM, Kurenkov2020VisuomotorMS, Zhang2018AMC} From the above example, since the apple (the target object) is obscured by a toothbrush, the robot must initially displace the toothbrush to successfully retrieve and deliver the apple. This task was previously resolved with a complicated POMDP \cite{zhang2021invigorate} integrated with multiple neural network modules specifically designed to learn the correct grasping order of objects. 

In Fig \ref{fig:1}, we demonstrate that even without specific training (zero-shot) \cite{zhang2023large, kojima2022large}, the language model proves capable of competently undertaking this task. This shows promising results for decision-making processes in robotic systems without the need for extensive, task-specific training.

\subsection{Disambiguation of Target Object}

Using LLMs to disambiguate the target object is the primary focus of this report. Initially, even with zero shots, the model is observed to surpass the performance capabilities of the enumerate method. It appears, considering Fig \ref{fig:2}, that for simple scenes with few objects and features, a large language model can successfully disambiguate to any target object. With each question, the number of potential target objects is halved (four cups to two cups with the first question, and two cups to one cup with the second).

    \begin{figure}[H]
    \centering
    \includegraphics[scale=0.5]{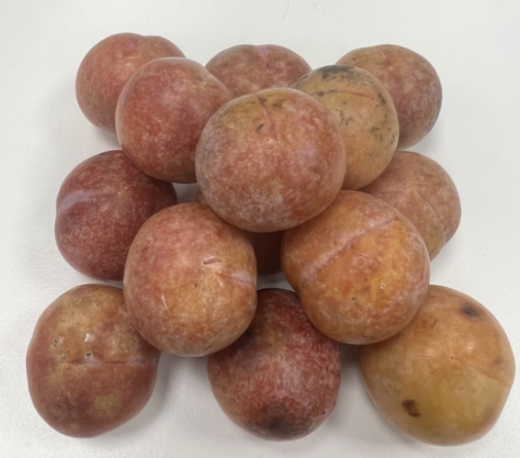}
    \caption{\textbf{Complex Scene Example. Fails with Zero-Shot Planning, but Succeeds with Few-Shot Prompting} ``There are 14 plums stacked in a pyramid on the table. On the bottom of the pyramid is a three by three square arrangement of 9 plums. The second layer rests on top of the bottom layer of 9 plums and consists of a two by two square arrangement of 4 plums. Finally on the top of the pyramid, there is one plum that rests on top of the 4 plums of the second layer.''}
    \label{fig:3}
\end{figure}

 \begin{figure*}[!t]
    \centering
    \includegraphics[scale=0.5]{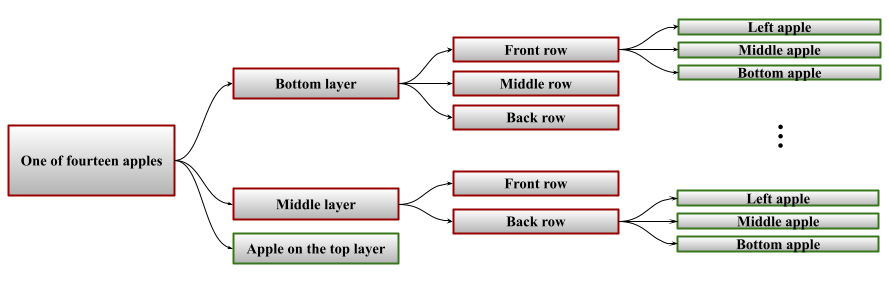}
    \caption{A visualization of a generated decision tree (originally in JSON format) from the trained model. The model’s initial disambiguation question seeks to identify the layer of the target object, thus reducing the options from any of the 14 apples to three distinct categories: apples located in the bottom layer, the middle layer, and the singular apple in the top layer. If the user indicates the bottom layer, the model will further refine its inquiry further by asking about the positioning within the layer (front row, middle row, or back row?). This process of narrowing down continues until the target object is unambiguously identified. }
    \label{fig:4}
\end{figure*}

Contrastingly, an enumerate method would have needed to pose four separate questions, each designed to clarify whether a specific object is indeed the target object. For example, instead of asking for the color of the target cup, INVIGORATE would ask if a specific cup is the desired one (“Do you want the cup on the left?”)\cite{zhang2021invigorate}.

With $n$ representing the total number of potential target objects, the previous INVIGORATE’s question-asking approach correlates with a worst-case scenario of $O(n)$, i.e., the number of questions increases linearly with the number of objects. Our model hopes to approach a logarithmic bound, reducing the worst-case number of questions asked, making it more efficient than its predecessor.

In Fig \ref{fig:2}, LLMs appear capable of disambiguating in simple scenes with few objects and zero-shot prompting, but when pushed further, it begins to show limitations in its capabilities. While LLMs demonstrate proficiency in disambiguating scenarios where abundant specified features are provided (for instance, when color, size, and classification are explicitly detailed in the scene description), they fall short in conjuring new features that are not already delineated in the language scene description. Take, for example, the scene in \ref{fig:3}, where there is a square pyramid of 14 plums. When the user requests a plum, an ideal action plan would deduce that the layer of the plum as well as the relative location within the layer could serve as a potential features to distinguish among the 14 plums. Although the second feature (relative-location within the layer) is not explicitly stated in the description, it can be reasonably inferred. Unfortunately, current pre-trained LLMs with zero-shot prompting aren’t equipped to make such deductions. The model’s response in this situation was: 

\begin{itemize}
    \item Action: <ask> <``Would you like a plum from the top of the pyramid, the bottom layer, or the middle layer?''>
    \item Reason:``There are multiple plums and we need to identify which one the human wants''
    \item Action: <deliver> <specified plum>
    \item Reason: ``After asking questions and receiving answers, we have identified the specific plum the human wants''
\end{itemize}

Although logically sound, this response shows the model's limitation in inferring unspecified features. In the scene description, layer was specified (bottom, middle, top), but relative location within each layer was not (top left, top middle, etc). The objective is to cultivate the use of spatial referring language, or other inferred features, without their explicit mention in the description. 

\section{Proposed Method}

To resolve this problem, we propose employing a few-shot prompt-engineering approach so that the LLM is capable of generating its own features. We refer the readers to Appendix for the full few-shot prompts. Post few-shot prompting, our method results in a model that is now capable of producing both a detailed action plan akin to those from previous examples, as well as a decision tree. This decision tree visually depicts how the disambiguating questions sequentially narrow down to a specific target object, demonstrating the process of honing in on the target object via one path down the tree.

\subsection{Example of Results}

Consider the scene in Fig \ref{fig:3}, the user's preference is to retrieve the plum on the far left corner of the bottom layer. The following illustrates the resulting interaction:

\begin{itemize}
    \item <ask> <``Would you like a plum from the bottom of the pyramid, the middle of the pyramid, or the top of the pyramid?''>
    \item User Response: ``the bottom layer''
    \item <ask> <``Would you like a plum from the \textbf{front row, the middle row, or the back row} of the bottom layer?''>
    \item User Response: ``the back row''
    \item <ask> <“Would you like the \textbf{``left plum, middle plum, or the right plum} from the back row of the bottom layer?''>
    \item User Response: ``left''
    \item <move away> <five plums on top of the left plum of the back row of the bottom layer>
    \item <deliver> <left plum of the back row of the bottom plum>
\end{itemize}

Note in bold are features that \textbf{were not once mentioned in the scene}. This illustrates a crucial advancement in the model’s ability to infer hidden information embedded in common sense knowledge exposed to the LLM during training, contributing to the method's overall decision-making competence. This generalizes to any target object of the user’s preference, and the action plans for disambiguation are summarized in Figure 4, the generated decision tree for disambiguation questions. 

\subsection{Summary of Significance}

The significance of this study lies in the model’s novel capability to self-generate features, expanding its comprehension and representation of a scene beyond explicit descriptions. This ability leverages the unique potential of pre-trained large language models that are exposed to massive amounts of data, an aspect that cannot be attained by previous methods, a POMDP for instance. 

Upon implementing our few-shot prompting approach, the model it's capability in producing comprehensive action plans capable of achieving efficient disambiguation in contrast with preceding methods, which were either inefficient or incomplete. This is further enhanced, as the model can now supplement the described scene with additional, self-derived features.  

\section{Experiments}

\begin{figure*}[]
    \centering
    \includegraphics[scale=0.4]{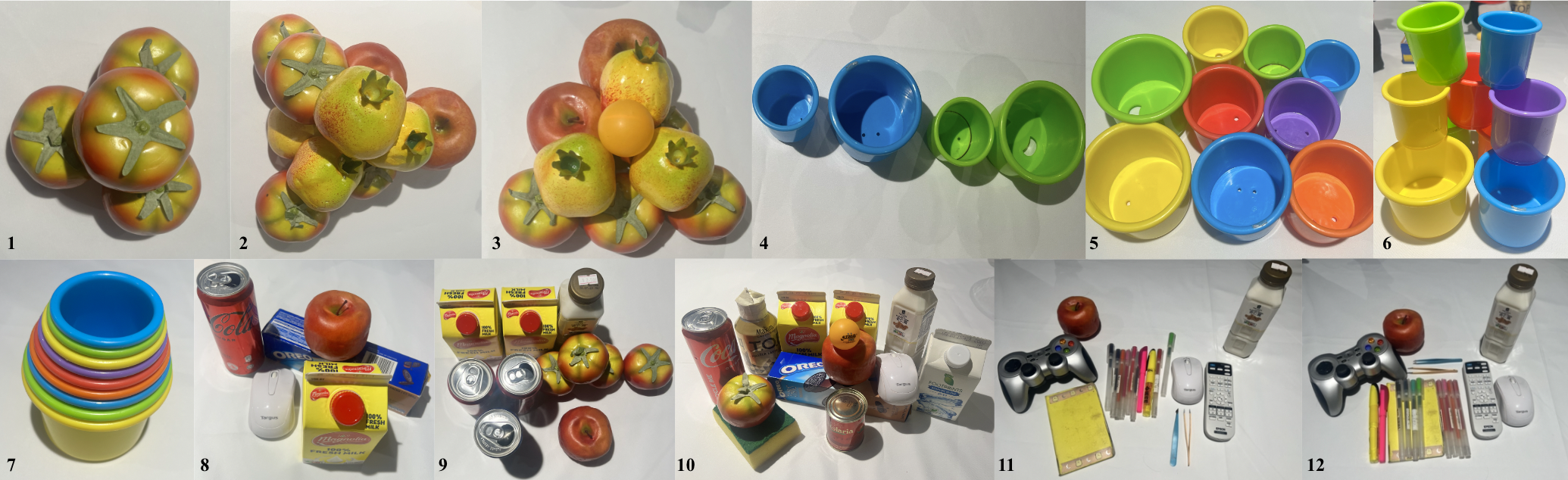}
    \caption{Image samples from the 12 scenes used for experiments.}
    \label{fig:5}
\end{figure*}

We evaluated our model’s performance and accuracy in diverse tabletop scenes, comparing it to four baseline methods:
\begin{itemize}
    \item \textbf{Optimal Split (ground truth)}: The optimal split of the decision tree possible given a scene, the objects, and their attributes.
    \item \textbf{Enumeration}: The method used in INVIGORATE \cite{zhang2021invigorate}. The robot randomly points to potential objects until the target object is verified, resulting in 100\% accuracy but extreme inefficiency. Given k potential objects, the average number of queries before identifying the target object is (k+1)/2.
    \item \textbf{Human Performance}: A human survey and experiment was conducted where the subject was given an image of the scene along with the language description of the scene, and the subject was then told there was a target object they were to identify, and they must ask sequential queries to disambiguate the target object. Again, accuracy was 100\% as humans do not stop asking questions until they have found the target object, but efficiency was surprisingly low due to the many extraneous questions and the human tendency to default to enumeration. For instance, sometimes a human would ask ``is the object edible?'', even when all the objects in the scene are edible. Or, after asking several questions, the subject will simply start pointing at each object in turn asking, ``is it this one?''
    \item \textbf{POMDP-ATTR} \cite{yang2022interactive}: The previous state-of-the-art model for disambiguation tasks. Has high efficiency in simpler scenes, but under-performs severely in more complex scenes, as it is limited to the basic colors and 9 locational vectors (top left, top middle, etc), which does not generalize to 3D object configurations (for instance if objects are stacked on top of each other). 
\end{itemize}

For our experiments, we created twelve distinct scenes, each with varied object configurations, feature combinations, and possible inquiries.

For each scene-inquiry pairing, we conducted three trials, producing a decision tree with all possible objects as the root node and target objects as the leaf nodes. We then recorded the average number of questions across all trials. 

\section{Results}

We use a decision tree to visualize and clearly define our results. The root node represents all potential target objects. Each leaf node represents a specific target object. Success is defined as a valid path from the root node to any leaf node on the decision tree. In other words, this involves successfully identifying the target object by asking questions while traversing down the tree. Our success rate is then calculated as the average ratio of successful target object identifications to the total number of objects across all trials. Using this approach, our model achieved a 95.79\% success rate, and the Figure 6 shows a table comparing the efficiency (quantified using the metric of number of questions posed), between baselines and our model.
\begin{figure}[h]
    \centering
    \includegraphics[scale=0.18]{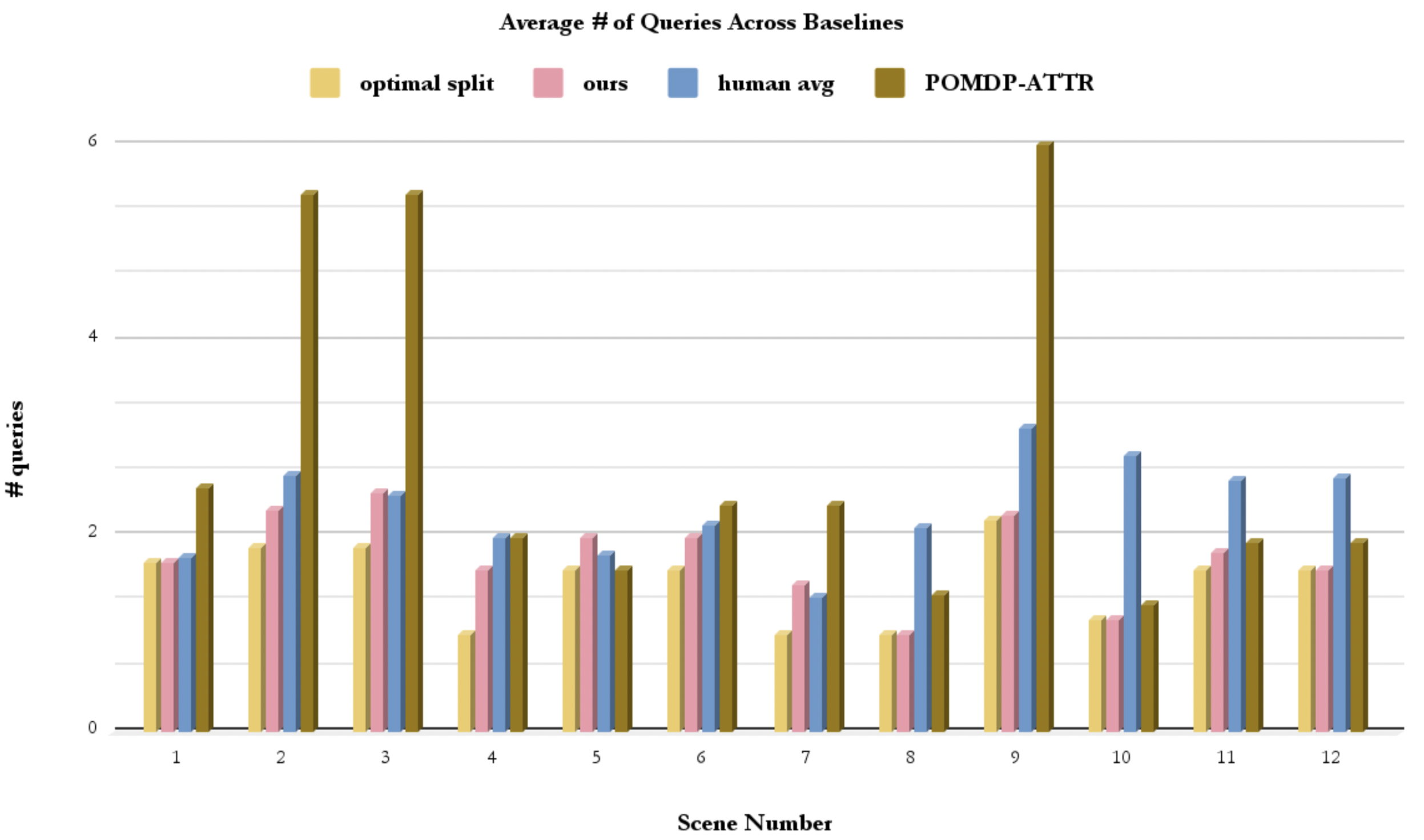}
    \caption{A comparative visualization of average query count in test scenes. Efficiency is represented by fewer queries. The optimal split (absolute minimum bound) is depicted in yellow (lightest in greyscale). Our model’s performance is depicted in pink (second-lightest in greyscale). Human performance and POMDP-ATTR’s performance are depicted in blue and brown (third-lightest and darkest) respectively. See Figure 5 for associated images of scenes 1-12.}
    \label{fig:enter-label}
\end{figure}
When given ambiguous queries, meaning any object in the given scene is a possible target object, our model is more efficient than human performance by 18.37\%, as human intuition does not always correlate to the most efficient splitting of features. Over all scenes, our model also surpasses POMDP-Attr by 26\%, mostly due to POMDP-Attr’s limitations in 3D tabletop object arrangements, where objects are stacked on top of one another. Ablating the scenes with stacked objects and only considering object arrangements flat on the table (Scenes 4, 5, and 12), our model asks 4.88\% fewer queries than POMDP-Attr. Table 1 provides a summary of disambiguation baseline comparison results.
\begin{table}[h]
\caption{Percentage Improvement in query count: our model vs. baselines}
\label{Disambiguation Comparison}
\begin{center}
\begin{tabular}{|c||c|}
\hline
Model & Improvement (\%)\\
\hline
Enumeration & 61.91\\
\hline
Human & 18.37\\
\hline
POMDP-Attr & 26.00\\
\hline
Optimal Split & -18.39\\
\hline
\end{tabular}
\end{center}
\end{table}
\section{Next Steps}

The most immediate next step would be to complete the visual portion of the pipeline. Currently, our model accepts natural language descriptions of scenes as input. A visual language model would need to be implemented to convert RGBD images of scenes into language descriptions.

\bibliography{bibliography.bib}

\begin{thebibliography}{10}
\providecommand{\url}[1]{#1}
\csname url@samestyle\endcsname
\providecommand{\newblock}{\relax}
\providecommand{\bibinfo}[2]{#2}
\providecommand{\BIBentrySTDinterwordspacing}{\spaceskip=0pt\relax}
\providecommand{\BIBentryALTinterwordstretchfactor}{4}
\providecommand{\BIBentryALTinterwordspacing}{\spaceskip=\fontdimen2\font plus
\BIBentryALTinterwordstretchfactor\fontdimen3\font minus
  \fontdimen4\font\relax}
\providecommand{\BIBforeignlanguage}[2]{{%
\expandafter\ifx\csname l@#1\endcsname\relax
\typeout{** WARNING: IEEEtran.bst: No hyphenation pattern has been}%
\typeout{** loaded for the language `#1'. Using the pattern for}%
\typeout{** the default language instead.}%
\else
\language=\csname l@#1\endcsname
\fi
#2}}
\providecommand{\BIBdecl}{\relax}
\BIBdecl

\bibitem{singh2023progprompt}
I.~Singh, V.~Blukis, A.~Mousavian, A.~Goyal, D.~Xu, J.~Tremblay, D.~Fox,
  J.~Thomason, and A.~Garg, ``Progprompt: Generating situated robot task plans
  using large language models,'' in \emph{2023 IEEE International Conference on
  Robotics and Automation (ICRA)}.\hskip 1em plus 0.5em minus 0.4em\relax IEEE,
  2023, pp. 11\,523--11\,530.

\bibitem{li2022pre}
S.~Li, X.~Puig, C.~Paxton, Y.~Du, C.~Wang, L.~Fan, T.~Chen, D.-A. Huang,
  E.~Aky{\"u}rek, A.~Anandkumar \emph{et~al.}, ``Pre-trained language models
  for interactive decision-making,'' \emph{Advances in Neural Information
  Processing Systems}, vol.~35, pp. 31\,199--31\,212, 2022.

\bibitem{chen2023introspective}
L.~Chen, L.~Wang, H.~Dong, Y.~Du, J.~Yan, F.~Yang, S.~Li, P.~Zhao, S.~Qin,
  S.~Rajmohan \emph{et~al.}, ``Introspective tips: Large language model for
  in-context decision making,'' \emph{arXiv preprint arXiv:2305.11598}, 2023.

\bibitem{hagendorff2023human}
T.~Hagendorff and S.~Fabi, ``Human-like intuitive behavior and reasoning biases
  emerged in language models--and disappeared in gpt-4,'' \emph{arXiv preprint
  arXiv:2306.07622}, 2023.

\bibitem{wang2019rat}
B.~Wang, R.~Shin, X.~Liu, O.~Polozov, and M.~Richardson, ``Rat-sql:
  Relation-aware schema encoding and linking for text-to-sql parsers,''
  \emph{arXiv preprint arXiv:1911.04942}, 2019.

\bibitem{kitaev2018multilingual}
N.~Kitaev, S.~Cao, and D.~Klein, ``Multilingual constituency parsing with
  self-attention and pre-training,'' \emph{arXiv preprint arXiv:1812.11760},
  2018.

\bibitem{zhu2020incorporating}
J.~Zhu, Y.~Xia, L.~Wu, D.~He, T.~Qin, W.~Zhou, H.~Li, and T.-Y. Liu,
  ``Incorporating bert into neural machine translation,'' \emph{arXiv preprint
  arXiv:2002.06823}, 2020.

\bibitem{brown2020language}
T.~Brown, B.~Mann, N.~Ryder, M.~Subbiah, J.~D. Kaplan, P.~Dhariwal,
  A.~Neelakantan, P.~Shyam, G.~Sastry, A.~Askell \emph{et~al.}, ``Language
  models are few-shot learners,'' \emph{Advances in neural information
  processing systems}, vol.~33, pp. 1877--1901, 2020.

\bibitem{keskar2019ctrl}
N.~S. Keskar, B.~McCann, L.~R. Varshney, C.~Xiong, and R.~Socher, ``Ctrl: A
  conditional transformer language model for controllable generation,''
  \emph{arXiv preprint arXiv:1909.05858}, 2019.

\bibitem{zhang2021invigorate}
H.~Zhang, Y.~Lu, C.~Yu, D.~Hsu, X.~La, and N.~Zheng, ``Invigorate: Interactive
  visual grounding and grasping in clutter,'' \emph{arXiv preprint
  arXiv:2108.11092}, 2021.

\bibitem{yang2022interactive}
Y.~Yang, X.~Lou, and C.~Choi, ``Interactive robotic grasping with
  attribute-guided disambiguation,'' in \emph{2022 International Conference on
  Robotics and Automation (ICRA)}.\hskip 1em plus 0.5em minus 0.4em\relax IEEE,
  2022, pp. 8914--8920.

\bibitem{li2017learning}
Y.~Li, C.~Huang, X.~Tang, and C.~Change~Loy, ``Learning to disambiguate by
  asking discriminative questions,'' in \emph{Proceedings of the IEEE
  International Conference on Computer Vision}, 2017, pp. 3419--3428.

\bibitem{shridhar2020ingress}
M.~Shridhar, D.~Mittal, and D.~Hsu, ``Ingress: Interactive visual grounding of
  referring expressions,'' \emph{The International Journal of Robotics
  Research}, vol.~39, no. 2-3, pp. 217--232, 2020.

\bibitem{lutkebohle2009curious}
I.~Lutkebohle, J.~Peltason, L.~Schillingmann, B.~Wrede, S.~Wachsmuth,
  C.~Elbrechter, and R.~Haschke, ``The curious robot-structuring interactive
  robot learning,'' in \emph{2009 IEEE International Conference on Robotics and
  Automation}.\hskip 1em plus 0.5em minus 0.4em\relax IEEE, 2009, pp.
  4156--4162.

\bibitem{mai2023llm}
J.~Mai, J.~Chen, B.~Li, G.~Qian, M.~Elhoseiny, and B.~Ghanem, ``Llm as a
  robotic brain: Unifying egocentric memory and control,'' \emph{arXiv preprint
  arXiv:2304.09349}, 2023.

\bibitem{ding2023task}
Y.~Ding, X.~Zhang, C.~Paxton, and S.~Zhang, ``Task and motion planning with
  large language models for object rearrangement,'' \emph{arXiv preprint
  arXiv:2303.06247}, 2023.

\bibitem{inoue2022prompter}
Y.~Inoue and H.~Ohashi, ``Prompter: Utilizing large language model prompting
  for a data efficient embodied instruction following,'' \emph{arXiv preprint
  arXiv:2211.03267}, 2022.

\bibitem{liu2023llm+}
B.~Liu, Y.~Jiang, X.~Zhang, Q.~Liu, S.~Zhang, J.~Biswas, and P.~Stone, ``Llm+
  p: Empowering large language models with optimal planning proficiency,''
  \emph{arXiv preprint arXiv:2304.11477}, 2023.

\bibitem{tellexll2013toward}
S.~Tellexll, P.~Thakerll, R.~Deitsl, D.~Simeonovl, T.~Kollar, and N.~Royl,
  ``Toward information theoretic human-robot dialog,'' \emph{Robotics}, p. 409,
  2013.

\bibitem{thomason2019improving}
J.~Thomason, A.~Padmakumar, J.~Sinapov, N.~Walker, Y.~Jiang, H.~Yedidsion,
  J.~Hart, P.~Stone, and R.~J. Mooney, ``Improving grounded natural language
  understanding through human-robot dialog,'' in \emph{2019 International
  Conference on Robotics and Automation (ICRA)}.\hskip 1em plus 0.5em minus
  0.4em\relax IEEE, 2019, pp. 6934--6941.

\bibitem{yang2021attribute}
Y.~Yang, Y.~Liu, H.~Liang, X.~Lou, and C.~Choi, ``Attribute-based robotic
  grasping with one-grasp adaptation,'' in \emph{2021 IEEE International
  Conference on Robotics and Automation (ICRA)}.\hskip 1em plus 0.5em minus
  0.4em\relax IEEE, 2021, pp. 6357--6363.

\bibitem{silver2022pddl}
T.~Silver, V.~Hariprasad, R.~S. Shuttleworth, N.~Kumar, T.~Lozano-P{\'e}rez,
  and L.~P. Kaelbling, ``Pddl planning with pretrained large language models,''
  in \emph{NeurIPS 2022 Foundation Models for Decision Making Workshop}, 2022.

\bibitem{yang2023harnessing}
J.~Yang, H.~Jin, R.~Tang, X.~Han, Q.~Feng, H.~Jiang, B.~Yin, and X.~Hu,
  ``Harnessing the power of llms in practice: A survey on chatgpt and beyond,''
  \emph{arXiv preprint arXiv:2304.13712}, 2023.

\bibitem{naveed2023comprehensive}
H.~Naveed, A.~U. Khan, S.~Qiu, M.~Saqib, S.~Anwar, M.~Usman, N.~Barnes, and
  A.~Mian, ``A comprehensive overview of large language models,'' \emph{arXiv
  preprint arXiv:2307.06435}, 2023.

\bibitem{yin2023large}
Z.~Yin, Q.~Sun, Q.~Guo, J.~Wu, X.~Qiu, and X.~Huang, ``Do large language models
  know what they don't know?'' \emph{arXiv preprint arXiv:2305.18153}, 2023.

\bibitem{huang2022language}
W.~Huang, P.~Abbeel, D.~Pathak, and I.~Mordatch, ``Language models as zero-shot
  planners: Extracting actionable knowledge for embodied agents,'' in
  \emph{International Conference on Machine Learning}.\hskip 1em plus 0.5em
  minus 0.4em\relax PMLR, 2022, pp. 9118--9147.

\bibitem{zhang2023large}
B.~Zhang and H.~Soh, ``Large language models as zero-shot human models for
  human-robot interaction,'' \emph{arXiv preprint arXiv:2303.03548}, 2023.

\bibitem{kojima2022large}
T.~Kojima, S.~S. Gu, M.~Reid, Y.~Matsuo, and Y.~Iwasawa, ``Large language
  models are zero-shot reasoners,'' \emph{Advances in neural information
  processing systems}, vol.~35, pp. 22\,199--22\,213, 2022.

\bibitem{wu2023tidybot}
J.~Wu, R.~Antonova, A.~Kan, M.~Lepert, A.~Zeng, S.~Song, J.~Bohg,
  S.~Rusinkiewicz, and T.~Funkhouser, ``Tidybot: Personalized robot assistance
  with large language models,'' \emph{arXiv preprint arXiv:2305.05658}, 2023.

\bibitem{Danielczuk2019MechanicalSM}
\BIBentryALTinterwordspacing
M.~Danielczuk, A.~Kurenkov, A.~Balakrishna, M.~Matl, D.~Wang,
  R.~Mart{\'i}n-Mart{\'i}n, A.~Garg, S.~Savarese, and K.~Goldberg, ``Mechanical
  search: Multi-step retrieval of a target object occluded by clutter,''
  \emph{2019 International Conference on Robotics and Automation (ICRA)}, pp.
  1614--1621, 2019. [Online]. Available:
  \url{https://api.semanticscholar.org/CorpusID:67877029}
\BIBentrySTDinterwordspacing

\bibitem{Kurenkov2020VisuomotorMS}
\BIBentryALTinterwordspacing
A.~Kurenkov, J.~C. Taglic, R.~Kulkarni, M.~Dominguez-Kuhne, A.~Garg,
  R.~Mart'in-Mart'in, and S.~Savarese, ``Visuomotor mechanical search: Learning
  to retrieve target objects in clutter,'' \emph{2020 IEEE/RSJ International
  Conference on Intelligent Robots and Systems (IROS)}, pp. 8408--8414, 2020.
  [Online]. Available: \url{https://api.semanticscholar.org/CorpusID:221135795}
\BIBentrySTDinterwordspacing

\bibitem{Zhang2018AMC}
\BIBentryALTinterwordspacing
H.~Zhang, X.~Lan, S.~Bai, L.~Wan, C.~Yang, and N.~Zheng, ``A multi-task
  convolutional neural network for autonomous robotic grasping in object
  stacking scenes,'' \emph{2019 IEEE/RSJ International Conference on
  Intelligent Robots and Systems (IROS)}, pp. 6435--6442, 2018. [Online].
  Available: \url{https://api.semanticscholar.org/CorpusID:67855510}
\BIBentrySTDinterwordspacing

\end{thebibliography}
\bibliographystyle{IEEEtran}

\newpage
\onecolumn
\appendix

\section*{Zero-Shot Prompt}
The following details the complete zero-shot prompt used to illustrate the successes and failures of zero-shot prompt engineering, as described in Section III, Part C. 
\begin{lstlisting}
      You are helping a household robot.
      The robot's task is, given a pile of items on a table and a specific inquiry, to retrieve an item and give it to the human.
      Initially, there might be multiple items that satisfy the inquiry. If this is the case, the robot must ask simple disambiguation questions 
      until it is sure that the target object is desired by the human. 
      The most important requirement is that the final target object must be ONE specific object in the scene with no ambiguity (pay attention to the total number of objects to ensure correct disambiguation).
      Your task is to give the robot direct instructions to retrieve a specific item.

      For each step, you can either ask a question, move away an object, or deliver one.
      Your action step-wise action format should be as follows: <action> <object/question> 
      The <action> can be <move away> or <deliver> or <ask>
      For example, if you want the robot to move the toothbrush away from the apple, your output should be <move away> <toothbrush>
      If you want the robot to deliver the toothbrush, your output should be <deliver> <toothbrush>
      If you want to clarify the color of the desired toothbrush, your output can be <ask> <do you mean the blue toothbrush?>
      Every time you choose to <ask>, specify a specific descriptor you wish to disambiguate. (location, relative location, size, etc) Note that the target object you finally 
      deliver should be a precise one. In other words, it should not be ambiguous, and you should continue to ask questions until there is only 
      ONE specific object (it can not be one of identical objects). 

      In asking disambiguation questions, note that certain descriptors might not have been previously specified in the given scene. 
      For example, if the given scene is "20 identical objects in a line", and the inquiry is "bring me an object", and that is all you are given,
      one example direction is <ask> <Which object from the left would you like? The one on the very left, second from the left, ... or the one on the right?>

      Your output should consist of two syntactically correct and verified JSON file versions of the following template (Ensure that the opening and clothing paranethesis and brackets match):

      Action Planner:
      {
        target object: <object>,
        reason: <reason>,
        direction: <action> <object>,
        reason: <reason>,
        direction: <action> <object>,
        reason: <reason>,
        ...
      }
      Decision Tree:
      {
        "possible objects": [
          "possible object",
          {
            "more possible objects": [
              "possible object",
              "possible object"
            ] 
          }
        ]
      }
\end{lstlisting}

\section*{Example for Few-shot prompting}
\label{sec:few-shotprompt}

Here, we detail one example in the few-shot prompting used in our few-shot prompt for querying the LLM to disambiguate the target objects. Note that this shot was included into the prompt as an addition to the zero-shot prompt.

\begin{lstlisting}
      For example, if the scene contains the following:
      "there is a toothbrush on top of an apple, and two chocolate bars side by side next to the toothbrush and apple."
      with the inquiry being: "bring me something to eat",

      An example of a possible response is: 

      Action Planner:
      {
        target object: <apple> or <chocolate bar>,
        reason: <An apple and a chocolate bar are both objects you can eat.>,
        direction: <ask> <Would you like an apple or a chocolate bar?>,
        reason: <I do not know whether the user wants an apple or a chocolate bar, as both are things you can eat.>,
        options: [
          <apple>: {
            target object: <apple>,
            reason: <The user clarified that they would like an apple>,
            direction: <move away> <toothbrush>,
            reason: <The toothbrush is in the way because it is on top of the apple.>,
            direction: <deliver> <apple>,
            reason: <There is nothing blocking the apple, and an apple is the target object.>
          },
          <chocolate bar>: {
            target object: <left chocolate bar> or <right chocolate bar>,
            reason: <There are two chocolate bars to choose from>,
            direction: <ask> <Would you like the chocolate bar on the left or the chocolate bar on the right?>,
            reason: <I do not know which chocolate bar the user would like>,
            options: [
              <left chocolate bar>: {
                target object: <left chocolate bar>,
                reason: <The user clarified they would like the chocolate bar on the left>,
                direction: <deliver> <left chocolate bar>,
                reason: <There is nothing blocking the chocolate bar, and the left one is the target object>
              },
              <right chocolate bar>: {
                target object: <right chocolate bar>,
                reason: <The user clarified they would like the chocolate bar on the right>,
                direction: <deliver> <right chocolate bar>,
                reason: <There is nothing blocking the chocolate bar, and the right one is the target object>
              }
            ]
          }
        ]
      }
      Decision Tree:
      {
        "something to eat": [
          "apple",
          {
            "chocolate bar": [
              "right chocolate bar",
              "left chocolate bar"
            ] 
          }
        ]
      }
\end{lstlisting}

\addtolength{\textheight}{-12cm}   




\end{document}